\documentclass{article}
\pdfoutput=1
\usepackage{spconf,amsmath,graphicx,epsfig}
\usepackage{pdfpages}
\usepackage{graphicx}
\usepackage{enumerate}
\usepackage{multirow}
\usepackage{caption}
\usepackage{subcaption}
\usepackage{url}

\renewcommand{\l}{\boldsymbol{l}}

\newcommand{\imageid}{001195}
\newcommand{\alphavariant}{1}
\newcommand{\netcutoff}{1}

\title{
Controlling Explanatory Heatmap Resolution and Semantics via Decomposition Depth
}
\name{Sebastian Bach$^1$, Alexander Binder$^{2,\ast}$, Klaus-Robert M\"uller$^{3,4,\ast}$ {\rm{and}} Wojciech Samek$^{1,\ast}$}

\address{$^1$ Fraunhofer Heinrich Hertz Institute, Einsteinufer 37, 10587 Berlin, Germany\\
$^2$ Singapore University of Technology (SUTD), 8 Somapah Road, Singapore 487372, Singapore\\
$^3$ Berlin Institute of Technology (TU Berlin), Stra{\ss}e des 17. Juni 135, 10623 Berlin, Germany\\
$^4$ Korea University, 145 Anam-ro, Seongbuk-gu, Seoul, 02841, Korea\\
$^\ast$ Member, IEEE
}

\begin{document}
\maketitle
\begin{abstract}

We present an application of the Layer-wise Relevance Propagation (LRP) algorithm to state of the art deep convolutional neural networks and Fisher Vector classifiers to compare the image perception and prediction strategies of both classifiers with the use of visualized heatmaps.
Layer-wise Relevance Propagation (LRP) is a method to compute scores for individual components of an input image, denoting their contribution to the prediction of the classifier for one particular test point.
We demonstrate the impact of different choices of decomposition cut-off points during the LRP-process, controlling the resolution and semantics of the heatmap on test images from the PASCAL VOC 2007 test data set.

\end{abstract}

%%%%%%%%%%%%%%%%%%%%%%%%%%%%%%%%%%%%%%%%%%%%%%%%%%%%%%%%%%%%
\section{Introduction} %use this command to squeeze page content
Nonlinear models play an integral part in many well-predicting machine learning algorithms. They include, for example, kernel machines, artifical neural networks, and other nonlinear mapping functions and feature space transformations e.g. during preprocessing steps. Many high-performing predictors consist of multiple layers of such mappings resulting in powerful predictive capabilities, which comes at the cost of an obfuscation of the decision making. Oftentimes, however, knowledge about \emph{how} a prediction comes to pass is as equally important as the confidence of the prediction, as it may e.g. help to identify the weaknesses of a classifier, the training data or reveal the policies followed by the predictor.

Quite recently, multiple endeavours have been made to gain insight into those black box classifiers, e.g. for neural network type classifiers by highlighting dominant filter activations \cite{DBLP:conf/eccv/ZeilerF14}, computing saliency maps visualizing local sensitivities \cite{DBLP:journals/corr/SimonyanVZ13} or the identification of support regions \cite{liu2012has} critical to the prediction for SVM \cite{cortes1995support} classifiers with max-pooling feature mapping and the explanation of hard feature mappings with HIK kernels \cite{uijlings2012visual}.

With Layer-wise Relevance Propagation \cite{bach2015pixel} (LRP), a principled approach applicable to a wide range of classifier architectures and problem domains has been introduced.
LRP is a method for explaining the output of a classifier wrt to the input data. Specifically, the method allows to generate an explanation of how the individual components of the input in their given state cause the evaluating model to arrive at its decision.
LRP assigns \emph{relevance scores} to each input (or intermediate representation) component, which can then be visualized as a \emph{heatmap}.
In \cite{bach2015analyzing}, the algorithm has been used to compare the perception of Deep Neural Network (DNN) classifiers and a state-of the art configuration of the improved Fisher Vector \cite{perronnin2010improving} (FV) classifier, demonstrating the method's general applicably to a wide range of classifier architectures.
However, due to limits in the transparencies of the feature extraction process which in general apply to Bag of Feature (BoF) classifiers, heuristic steps have been incorporated into the decomposition process \cite{bach2015analyzing, bach2015pixel,liu2012has}, leaving the decision-explaining heatmaps to appear much less sparse and at a lower resolution when compared to the output of LRP applied to DNNs. To alleviate this issue, we introduce the notion of a \emph{mapping influence cut-off point} with allows to control the degree of detail and semantics of a heatmap. A qualitative analysis on heatmaps at different resolutions computed for images of the PASCAL VOC 2007 \cite{everingham2008pascal} and both predictors is performed.

\section{Layer-wise Relevance Propagation}
The aim of LRP is to attribute shares of upper layer relevances $R^{(l+1)}_j$ to all components $i$ of the adjacent lower layer $l$, such that each component of $l$ receives a relevance score $R^{(l)}_i$ proportionally to its contribution to the output values at layer $l+1$ when performing a forward pass. In its simplest and most general formulation, this is realized via the local decomposition rule
\begin{align}
R_i^{(l)} = \sum\limits_j \frac{z_{ij}}{z_{j}} R_j^{(l+1)}
\label{eq:local_redistribution}
\end{align}
with $z_{ij}$ representing the outcome of a forward mapping operation from component $i$ of layer $l$ to component $j$ in layer $l+1$ and $z_j$ being the combined output at layer $l+1$. Note that the case $0/0$ is being treated as $0$. Starting at the predictor output with $R^{(l+1)} = f(x)$, this decomposition is then performed iteratively layer-by-layer under consideration of the classifier architecture until relevance scores at the input layer have been obtained. Several pertinent adaptions of Equation \ref{eq:local_redistribution} specific to different forward mappings have been discussed and evaluated in \cite{montavon2015explaining, bach2015pixel, bach2015analyzing}, with select variants relevant to the classifiers used within the scope of this paper being remarked in below Sections \ref{sec:dnn_lrp} and \ref{sec:bof_lrp}.

\subsection{LRP for Deep Neural Networks}
\label{sec:dnn_lrp}

Neural network type classifiers typically consist of a sequence of mapping layers
\begin{align}
z_{ij} = x_iw_{ij} ~;~ z_j = \sum_j z_{ij} + b_j ~;~ x_j = g(z_j)
\end{align}
where $x_i$ is the input, $w_{ij}$ and $b_j$ are the learned weights and bias terms and $g(\cdot)$ incorporates a non-linear activation or/and pooling function. This formulation of a network is in general enough to encompass a wide range or architectures, including convolution operations.
Equation \ref{eq:local_redistribution} and variants thereof are directly applicable to DNN classifiers, resulting in (sub)pixel-accurate\footnote{E.g. for DNNs receiving rgb color images as input, each color channel per pixel receives a relevance score. We use their sum as $R_p^{(1)}$ for vizualizations.}
%\footnote{E.g. in case of the input being a rgb color image, each pixel receives three relevance ratings -- one per color channel. For evaluation and visualization purposes we sum over the color channels to obtain scores $R_p^{(1)}$.}
relevance scores $R_p^{(1)}$ per input pixel $p$. For the ConvNet models with ReLu activation layers considered throughout this paper, we apply the $\alpha$/$\beta$-weighted decomposition formula with $\beta=1$, which has been identified to result in heatmaps best representing the classifier decision in \cite{samek2015evaluating}.

%we do in practice decompose positive and negative mapping contributions separately to prevent numerical issues due to (near-) zero divisions due to mutually cancelling mappings $z_{ij}$ when computing $z_j$:
%\begin{align*}
%R^{(l)}_i = \sum\limits_j \left( \alpha\frac{z_{ij}^{+}}{z_j^{+}} -\beta\frac{z_{ij}^{-}}{z_j^{-}} \right)R_j^{(l+1)}
%\end{align*}
%\textbf{FORMULA?BETA1?}
%Here, the denotations $+$ and $-$ identify the positive and negative parts of the mapping $z_{ij}$, $z_j^+=\sum_i z_{ij}^++b_j^+$ and analogously $z_j^-$. We constrain $\alpha + \beta = 1$ and set $\beta=1$.
%This alternative variant has been empirically evaluated to yield the \emph{best} heatmaps for representing the classifier decision\cite{all,applicable,papers} and further does adhere to the relevance conservation principle\footnote{formula or reference?}.

%The local redistribution rule described in  can be applied to arbitrary neural network layers with $z_{ij} = x_i^{(l)}w_{ij}^{(l,l+1)}$ being applicable to linear and fully connecting layers as well as convolution layers, where $x^{(l)}$ is the preactivated output of layer $l$ and $w$ a weight matrix connecting layer $l$ to its successor.
%The application of this rule in a backward pass produces a relevance map (heatmap) for pixels $\lbrace R^{(1)}_p \rbrace$ that satisfies the desired conservation property $\sum_p R_p^{(1)} = f(x)$, with $f(x)$ being equal to the initial relevance value at the output layer.

\subsection{LRP for Bag of Feature Classifiers}
\label{sec:bof_lrp}
The computations performed in the context of Bag of Feature methods fit well into the framework described by LRP. Similar to DNNs, BoF classifiers operate by executing consecutive layers of feature extraction, mappings and transformations relative to a visual vocabulary and pooling steps.
%, usually consisting of a feature extraction stage followed by a feature mapping, pooling step relative to a visual vocabulary,
The resulting vector description is then being fed into a support vector classifier for training and classification. LRP decomposes the prediction of such a pipeline in inverse direction, starting with $R^{(4)} = f(x)$. \cite{bach2015pixel} points out solutions for decomposing any kernel-based classification function to compute relevances $R^{(3)}_d$ for the dimensions $d$ of the vector representation of the input image.
Also, relevance scores $R^{(2)}_{\l}$ for all local descriptors $\l \in L$ as extracted from an input image can easily be computed for sum- and max-pooled (and anything in between) mappings.
In previous work, the notation $m(\l)_d$ has been used to describe the output of such a mapping function $m$ with input $\l$ to output dimension $d$, which corresponds to $z_{ij}$ (as $z_{\l d}$) in Equation \ref{eq:local_redistribution}.

Since feature extraction algorithms encode and compress the image within a the receptive field of a local descriptor, 
the relationship between single pixels and local descriptor dimensions -- or even visual prototypes -- is not inherently clear in general.
This is especially true for quantile-based feature descriptors.
For that reason \cite{bach2015pixel,bach2015analyzing} uniformly distribute a local feature's relevance score across all pixels within its receptive field to compute $R^{(1)}_p$ for each input pixel.
This leads to heatmaps appearing more coarse when compared to heatmaps produced from DNN classifiers (see Fig. \ref{fig:1}. and \cite{bach2015pixel,bach2015analyzing}), with heatmap granularity being limited by descriptor size. 

Within the scope of this paper we use a FV classifier to represent the class of BoF models, in a setup as described in \cite{chatfield2011devil}. The mappings of features $\l$ onto a FV representation may be of both positive and negative sign, potentially destabilizing the relevance decomposition. We therefore employ the following $\epsilon$-stabilized decomposition formula with $\epsilon=100$. This redistribution rule has been found to compute explanatory heatmaps matching the classifier decision well \cite{bach2015analyzing}.

%\begin{align*}
%R^{(2)}_{\l} = \sum\limits_{d \not\in Z(x)} \frac{z_{\l d}}{z_d + \epsilon\cdot sign(z_d)}R^{(3)}_d + \xi 
%\end{align*}
%with $Z(x)=\lbrace d | \forall \l~:~ z_{\l d}=0\rbrace$ and $\xi=\sum_{d\in Z(x)} R^{(3)}_d/|L|$. This redistribution rule has been found to compute explanatory heatmaps matching the classifier decision well \cite{bach2015analyzing}. The term $\xi$ redistributes relevance values from FV dimensions not reached by any mapped descriptor. This ensures the relevance conservation principle holds, aside from small amounts of relevance being absorbed by $\epsilon$.

\section{Controlling heatmap detail and semantic}
\label{sec:cut-off}
We introduce the notion of an \emph{mapping influence cut-off point}, describing the step from which on the forward mapping function of the classifier does not influence relevance propagation anymore and only the receptive field of the classifier does. For the DNN architecture as described in Section \ref{sec:dnn_lrp}, there is no such cut-off point being used in previous work, whereas for the BoF architecture as of Section \ref{sec:bof_lrp} the cut-off is located at the decomposition layer resulting in $R^{(1)}_p$ (for brevity, we say the cut-off is ``at $R^{(1)}_p$'') and has been chosen out of necessity in past work.
In this Section our aim is to explain how the choice of such a cut-off point may be voluntary for both considered predictor architectures by potentially increasing or decreasing the heatmap resolution for BoF models and DNNs respectively.

For neural network type classifiers, decreasing the resolution of LRP-computed heatmaps follows the procedure to compute scores $R^{(1)}_p$ for BoF classifiers. For ConvNets, instead of applying the local decomposition rule from Equation \ref{eq:local_redistribution} to the bottom-most convolution layer, relevance scores computed for the succeeding pooling layer are to be distributed uniformly across all of its inputs.
This is equivalent to applying Equation \ref{eq:local_redistribution} and then averaging lower layer relevances for each convolution operator -- or in general, choosing a cut-off point at a higher layer is equal to substituting any mappings  from that layer on with flat weights.
  Note that for simple fully connected layers, this approach will render the resulting lower layer relevances meaningless, since spatial structure -- as it is the case with convolution layers -- is not present and the resulting heatmap will be uniform. Here, the $w^2$-rule proposed in \cite{montavon2015explaining} might yield satisfactory results by distributing relevance values according to mapping weights only
%\footnote{should be equal to what Zeiler does, with full-image convolution maps. write that?}.

To bridge the gap from local feature relevance scores to pixel relevance scores with BoF classifiers, two obstacles need to be outmanoeuvred, namely (i) one has to be able to compute relevance scores for each dimension of a local descriptor $\l$ and (ii) the local feature dimensions need to be relatable to a grid of pixel coordinates.
In this work, we concentrate on the FV classification model as described in \cite{chatfield2011devil}, where the requirements to solve (i) are given fully and (ii) partially, as explained below.

(i) Computing relevances $R^{(2)}_{\l_i}$ for local feature all local feature dimensions $i$: To compute relevance scores at such a fine granularity, the forward mapping contribution of each feature dimension needs to be known (or at least its \emph{influence}. 
%A short discussion will follow below/
This, however, will not be subject of this work).
For the FV model considered here, this is the case, such that an exact relevance composition can be performed for each $\l_i$. For one, \cite{chatfield2011devil} compute projections %$\Psi_{\mu_k}(\l)$ and $\Psi_{\sigma_k}(\l)$ of $\l$
\begin{align}
\Psi_{\mu_k}(\l) & = \frac{1}{\sqrt{\pi_k}}\gamma_k(\l)\left( \frac{\l - \mu_k}{\sigma_k}\right) \\
\Psi_{\sigma_k}(\l) & = \frac{1}{\sqrt{2\pi_k}}\gamma_k(\l)\left( \frac{\left(\l - \mu_k\right)^2}{\sigma_k^2} - 1\right)
\end{align}
relating $\l$ to all $K$ components of a GMM $\lambda = \lbrace (\pi_k, \mu_k, \Sigma_k) \rbrace_{1..K}$ fit during training wrt to its 1st and 2nd moments, with $\l\in \mathrm{R}^D$ and also $\Psi_{\mu_k}(\l)$ and $\Psi_{\sigma_k}(\l)\in \mathrm{R}^D$. Further, the convariance matrices of the trained GMM have been constrained to be diagonal (e.g. $\forall k: \Sigma_k = diag(\sigma_k) $). We therefore know that each $\l_i$ corresponds to exactly one dimension in the mapping output space of $\Psi_{\mu_k}(\l)$ and $\Psi_{\sigma_k}(\l)$ for all $k$, which are concatenated to form the full FV representation $\Psi_{\lambda}(\l)$. For simplicity, suppose a function $d = \delta(i,\Psi_{{\lbrace\mu,\sigma\rbrace}_k})$, which computes for a local feature dimension $i$ and mapping of choice the output dimension $d$ of the FV representation. We compute relevance scores for each $\l_i$ as
\begin{align}
R^{(2)}_{\l_i}  = \sum\limits_{k=1}^K&\left( \frac{\Psi_{\mu_k}(\l)_i}{\sum_{\l' \in L}\Psi_{\mu_k}(\l')_i}R^{(3)}_{\delta(i,\Psi_{\mu_k})} \right.\\
				& + \left. \frac{\Psi_{\sigma_k}(\l)_i}{\sum_{\l' \in L}\Psi_{\sigma_k}(\l')_i}R^{(3)}_{\delta(i,\Psi_{\sigma_k})} \right)	.			
\end{align}
Note that in practice we do still apply the $\epsilon$-stabilized decomposition variant. Above decomposition describes the basic approach. Before mapping the local descriptors into the FV space, the reference model from \cite{chatfield2011devil} projects the initial $128$-dimensional SIFT \cite{lowe2004distinctive} features onto a $80$-dimensional subspace via mapping components computed with PCA \cite{pearson1901liii} during training. The now also $80$-dimensional $R^{(2)}_{\l}$ can easily be projected into the original SIFT space by applying Equation \ref{eq:local_redistribution}, since the appropriate operation in the forward pass realizes a linear projection. A numerical stabilization of the denominator was necessary for good results, and $\epsilon=100$ has been chosen as the best suited parameter after visual inspection. % \todo{fig?}.

(ii) Relating local feature dimensions to a spatial pixel grid: Many local feature types aggregate information extracted from an image area, such that the relation between feature dimension and pixel coordinate is lost in the process on the interaction between groups pixel values. An appropropriate example are quantile-based local descriptors, for the computation of which all scanned pixel values are of importance, yet only a select few are grouped into the final descriptor, maybe even in an interpolated manner.
However, knowledge about spatial bin placement and local feature geometry can be used whenever possible in order to intelligently merge $R^{(2)}_{\l_i}$ and assign relevance scores to pixels at a higher resolution.
The reference FV predictor uses SIFT descriptors at different sizes with $4\times 4$ spatial bins, with each capturing a histogram of gradient magnitudes in $8$ directions. We use knowledge about the feature geometry and indexing \cite{vedaldi2010vlfeat,vlphow} to evenly distribute the summed relevances corresponding to each spatial bin evenly over the covered pixels, thus increasing the resolution of each SIFT feature's relevance feedback $16$-fold.

\section{Results}

We compute heatmap explanations for the FV model configured after \cite{chatfield2011devil} and trained for \cite{bach2015analyzing} and the BVLC reference model from the Caffe package \cite{jia2014caffe} which has been retrained for the 20 classes of PASCAL VOC 2007 \cite{everingham2008pascal}. Classification results for both models in average precision (AP) are listed in Table \ref{tab:fv_cnn_mAP}.
\begin{table}
{
\small
\centering
\begin{tabular}{|c||c|c|c|c|c|c|c|}
\hline
&    { \bf aer} & { \bf bic} & { \bf bir} & { \bf boat} & { \bf bot} & { \bf bus} & { \bf car}\\ 
{\bf F} & 79.1 & 66.4 & 45.9 & 70.9 & 27.6 & 69.7 & 81.0 \\ 
{\bf N} & 88.1 & 79.7 & 80.8 & 77.2 & 35.5 & 72.7 & 86.3 \\\hline 
&    { \bf cat} & { \bf cha} & { \bf cow} & { \bf din} & { \bf dog} & { \bf hor} & { \bf mot}\\ 
{\bf F} & 59.9 & 51.9 & 47.6 & 58.1 & 42.3 & 80.5 & 69.3 \\ 
{\bf N} & 81.1 & 51.0 & 61.1 & 64.6 & 76.2 & 81.6 & 79.3 \\ \hline 
&    { \bf per} & { \bf pot} & { \bf she} & { \bf sof} & { \bf tra} & { \bf tvm} & { \bf mAP}\\ 
{\bf F} & 85.1 & 28.6 & 49.6 & 49.3 & 82.7 & 54.3 & 60.0 \\ 
{\bf N} & 92.4 & 50.0 & 74.0 & 49.5 & 87.1 & 67.1 & 72.1 \\ \hline
\end{tabular}
}
%\vspace{10pt}
\caption{Prediction performance of the \textbf{F}V and DN\textbf{N} classifiers used in this paper, in average precision (AP) per class and in percent.}
\label{tab:fv_cnn_mAP}
\end{table}

\begin{figure}[th!]
\centering
\vspace{2mm}
\renewcommand{\imageid}{000801} 
\includegraphics[height=0.655in]{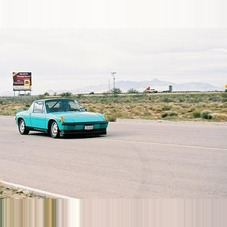}
\includegraphics[height=0.655in]{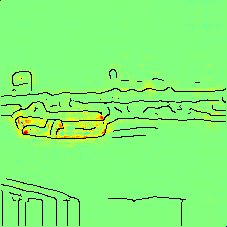} 
\includegraphics[height=0.655in]{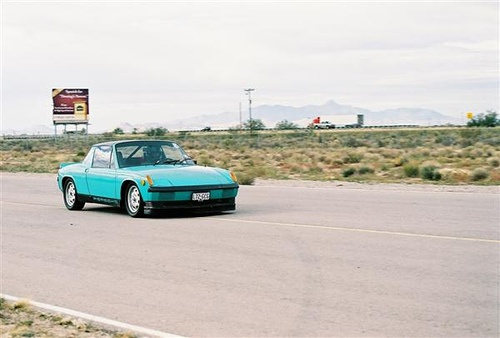}
\includegraphics[height=0.655in]{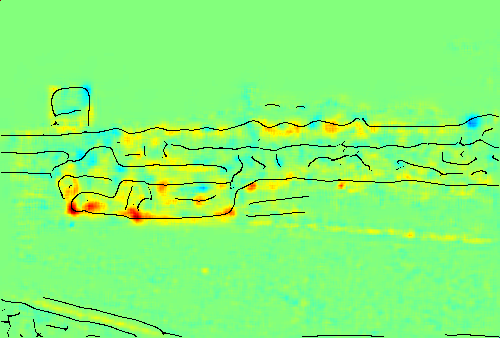}  \\
\includegraphics[height=0.655in]{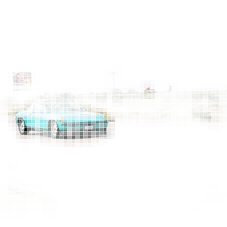}
\includegraphics[height=0.655in]{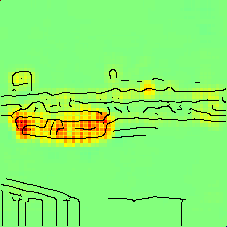}
\includegraphics[height=0.655in]{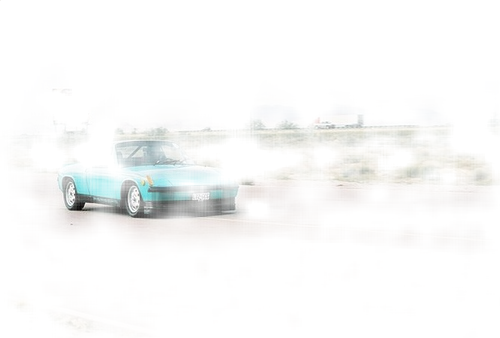}
\includegraphics[height=0.655in]{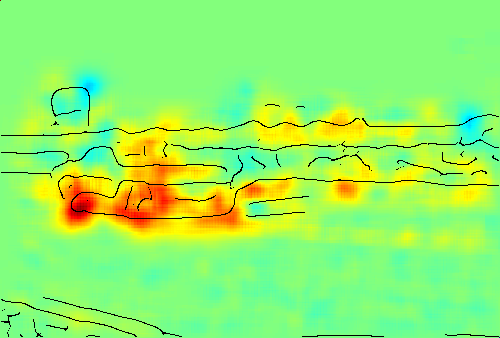}
\\
\vspace{2mm}
\renewcommand{\imageid}{001489}
\includegraphics[height=0.655in]{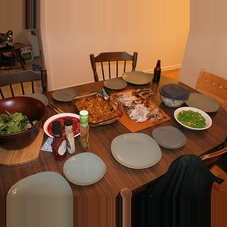}
\includegraphics[height=0.655in]{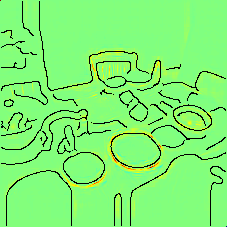} 
\includegraphics[height=0.655in]{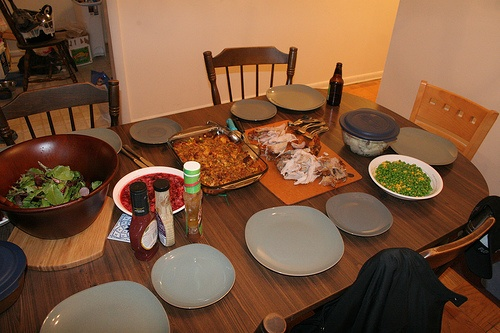}
\includegraphics[height=0.655in]{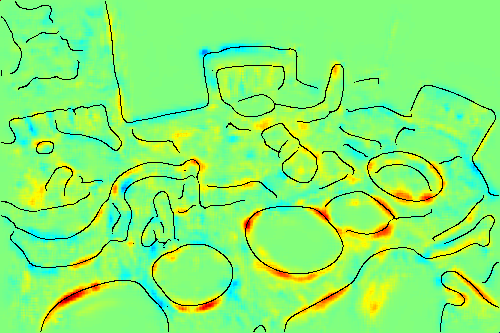}  \\
\includegraphics[height=0.655in]{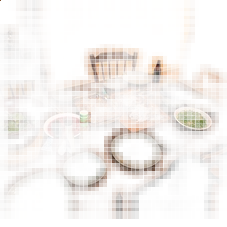}
\includegraphics[height=0.655in]{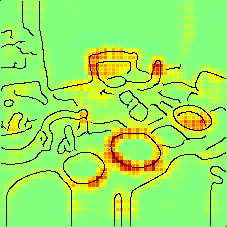}
\includegraphics[height=0.655in]{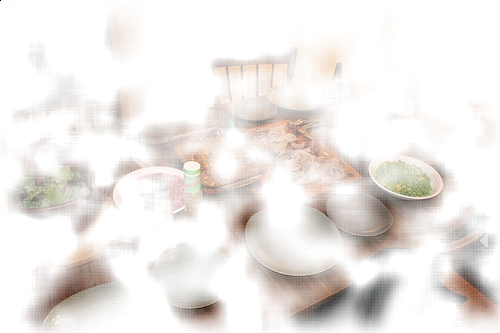}
\includegraphics[height=0.655in]{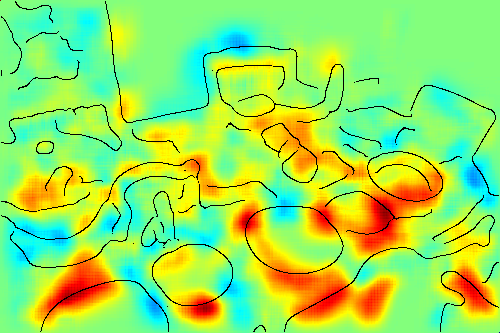}
\\
\vspace{2mm}
\renewcommand{\imageid}{001189}
\includegraphics[height=0.94in]{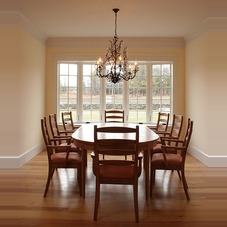}
\includegraphics[height=0.94in]{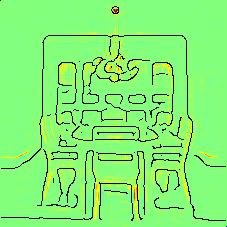} 
\includegraphics[height=0.94in]{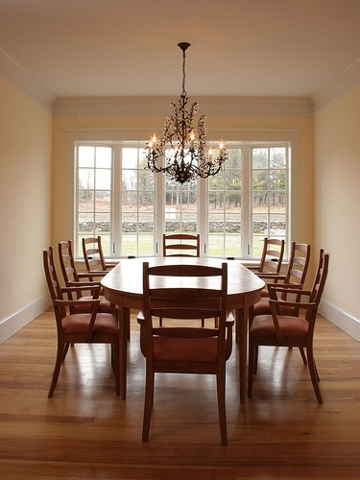}
\includegraphics[height=0.94in]{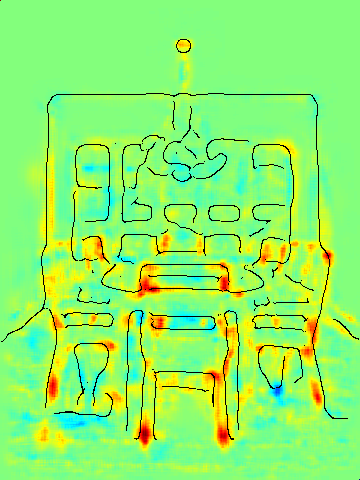}  \\
\includegraphics[height=0.94in]{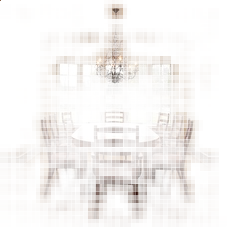}
\includegraphics[height=0.94in]{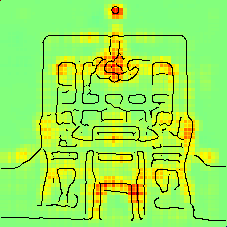}
\includegraphics[height=0.94in]{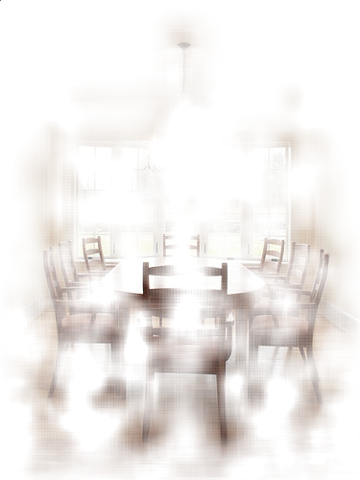}
\includegraphics[height=0.94in]{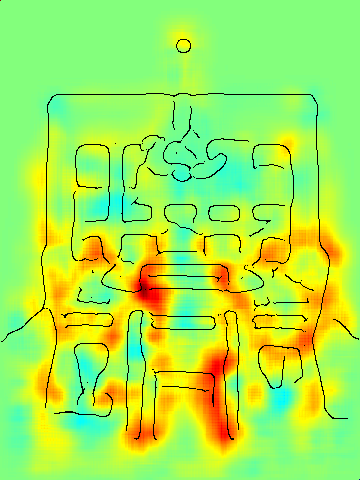}
\\
\vspace{2mm}
\renewcommand{\imageid}{001195}
\includegraphics[height=0.7in]{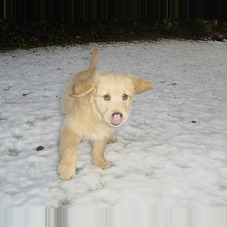}
\includegraphics[height=0.7in]{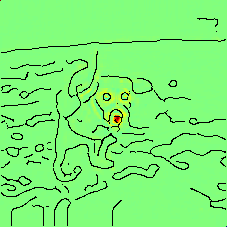}
\includegraphics[height=0.7in]{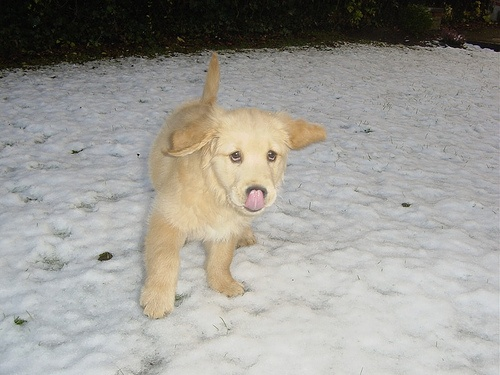}
\includegraphics[height=0.7in]{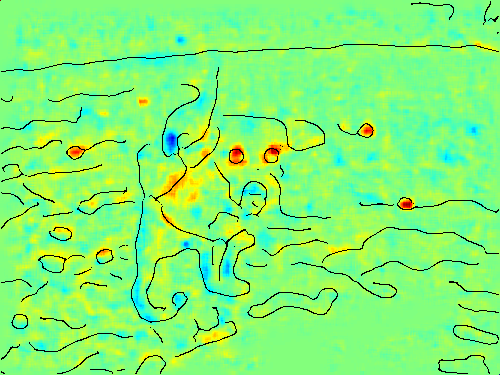}  \\
\includegraphics[height=0.7in]{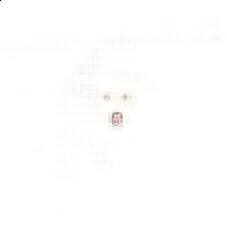}
\includegraphics[height=0.7in]{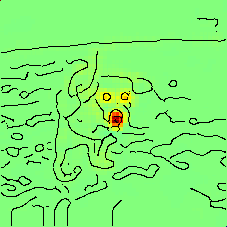}
\includegraphics[height=0.7in]{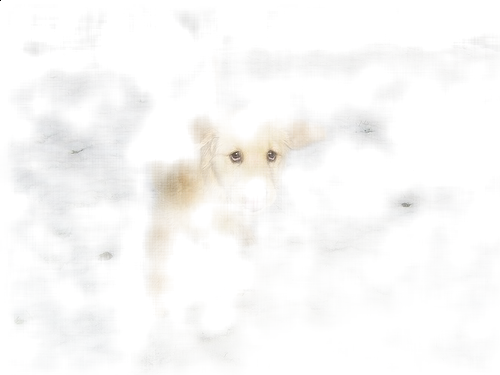}
\includegraphics[height=0.7in]{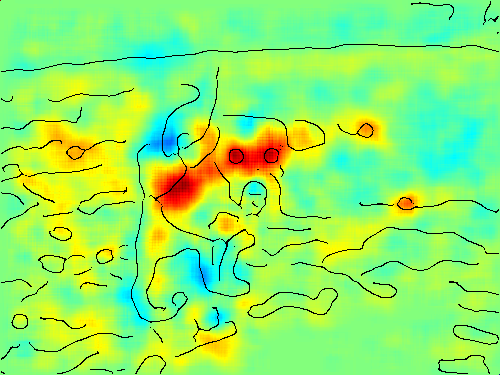}
\\
\vspace{2mm}
\renewcommand{\imageid}{001562}
\includegraphics[height=0.7in]{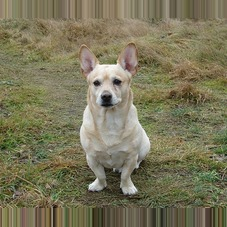}
\includegraphics[height=0.7in]{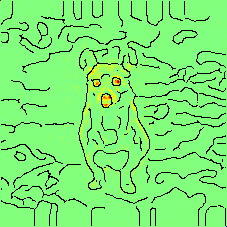} 
\includegraphics[height=0.7in]{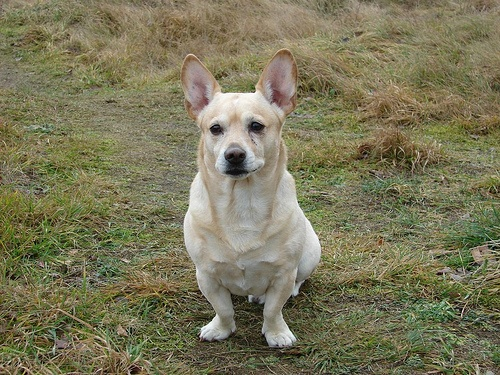}
\includegraphics[height=0.7in]{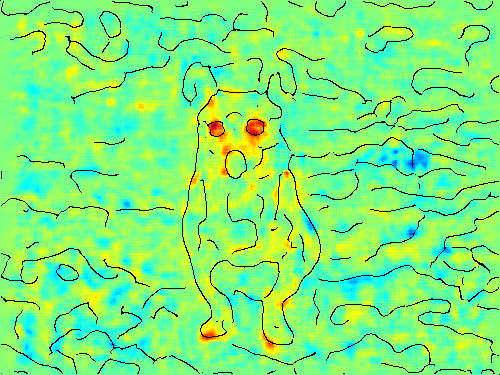}  \\
\includegraphics[height=0.7in]{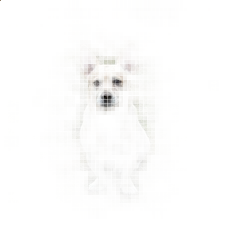}
\includegraphics[height=0.7in]{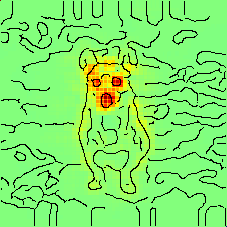}
\includegraphics[height=0.7in]{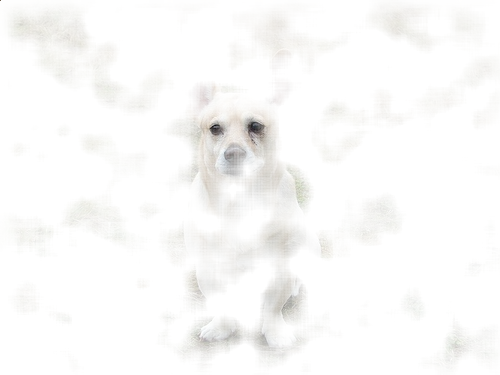}
\includegraphics[height=0.7in]{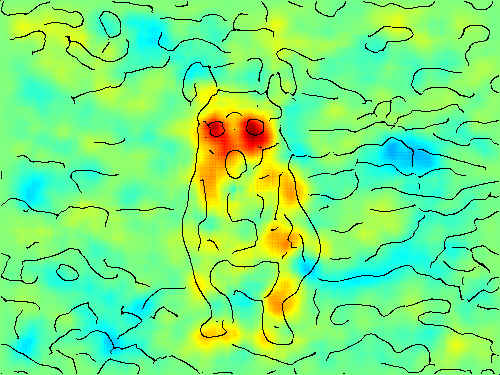}
\caption{\small Each two rows of images shows images corresponding to the CNN (left four images) and FV (right four) models. In clock-wise order: ({\tiny$^\nwarrow$}) Input image, ({\tiny$^\nearrow$}) high resolution heatmap without cut-off point, ({\tiny$\searrow$}) low resolution heatmap with cut-off at $R^{(1)}_p$ and ({\tiny$\swarrow$}) the same heatmap added as $\alpha$-channel to the input. Green heatmap areas are rated neutral to the classifier prediction, yellow to dark red hues indicate a positive contribution to the target class with image areas marked with blue color receiving negative relevance ratings.}
\label{fig:1}
\end{figure}

For both models, we compute heatmaps with cut-off points
at $R^{(1)}$
%above pixel level (FV) and at the $2$nd convolutional layer (DNN, due to the depth of the net)
and no cut-off (as far as possible for FV. See Section \ref{sec:cut-off}) and show results in Figure \ref{fig:1}.
We observe that heatmaps without cut-off are still sparser for DNNs due to the limits set to the FV classifier, despite images being fed into the network at a lower resolution.
Nonetheless, the higher resolution heatmaps for the FV classifier allow for a better understanding of the classifier decision when compared to the heatmaps with cut-off at $R^{(1)}$. In the example for class ``chair'' the high resolution relevances demonstrate that the FV classifier mostly uses the object structure itself for classification, which was difficult to read for the lower resolution heatmaps.
Both models seem to follow similar higher level strategies for most object classes, e.g. with the bottom half of cars, (also wheels of the bicycles, faces and clothing of people) or the faces of dogs capturing the models' focus of attention.
Both classifiers strongly react to defining aspects of the object, while still focussing on most of it as can be reasoned from the input images with the lower resolution heatmap controlling the alpha channel, visualizing the respective model's focus of attention.
We also observe that both classifiers seem to prefer the use of hard edges (e.g. cutlery for class table) for detection as the result of training (DNN) or due to design (FV with SIFT), rather than texture.
The more deep and complex DNN classifier seems to be in general more adept at abstracting object appearances and is therefore less prone to misleading noise information while using more sophisticated rules for prediction. For the image showing a dog, for example, the FV model reacts strongly to eye-like black pebbles in the snow, next to the dog's eyes themselves, whereas the network model is reacting to a structure resembling a dog's face (nose below eyes). However, this increased concentration of the DNN on higher structural information includes regularly co-appearing image features into the detection process, whereas the FV model tends to concentrate on simpler rules. Note the chandelier for class ``chair'' in living room scenes.
%, or the bicycle with its rider.

\section{Conclusion}

In this work we introduce the notion of a \emph{mapping influence cut-off point}, which allows to control the resolution of the computed heatmap and simultaneously its semantics. We have compared heatmap explanations for state-of-the art DNN and FV models at different degrees of detail. While (very) high resolution relevance maps provide information about the kind of visual cues a classifier has been conditioned on a very local level, lower resolution heatmaps shed light to the the classifier's focus of attention by directly mapping the relevance scores assigned to its receptive fields onto pixel level. By voluntarily choosing a mapping influence cut-off point, we are able to explain decisions wrt a desired aspect of the perception of the model.
In combination with image-wise predictions, heatmaps computed at different degrees of detail open up the possibility of automated assistance in different problem settings, e.g. in localizing target content upon detection, and then highlighting characteristic features.  We see possible applications for instance in the medical domain, where the screening of large images of stained tissue by a medical expert might be assisted with low resolution heatmaps for malignant tissue localization and a high resolution heatmap verifying the decision of the classifier.

\clearpage

\subsection*{Acknowledgment}
This work was supported by the German Ministry for Education and Research as Berlin Big Data Center BBDC (01IS14013A), the Deutsche Forschungsgesellschaft (MU 987/19-1) and the Brain Korea 21 Plus Program through the National Research Foundation of Korea funded by the Ministry of Education. AB acknowledges support by the SUTD Startup grant. Correspondence to KRM and WS.

%
%%%%%%%%%%%%%%%%%%%%%%%%%%%%%%%%%%%%%%%%%%%%%%%%%%%%%%%%%%%%
\bibliographystyle{IEEEbib}
\bibliography{icip2016}

\end{document}